# NUMERICAL SENSITIVITY AND EFFICIENCY IN THE TREATMENT OF EPISTEMIC AND ALEATORY UNCERTAINTY

E. Chojnacki *, J. Baccou, S. Destercke
*Institut de Radioprotection et de Sûreté Nucléaire (IRSN), DPAM, SEMIC, LIMSI, Cadarache, France*
*eric.chojnacki@irsn.fr*

## 1. Introduction

Nowadays, the need to treat both epistemic and aleatory uncertainty in a unified framework is well recognized [1]. One method to do so is to mix probabilistic convolution (for aleatory uncertainty) and fuzzy calculus (for epistemic uncertainty). Existing propositions either concern simple models [2] or are computationally very costly [3] (a luxury not always affordable, especially in nuclear safety, where models can be very complex). Here, we propose a numerical treatment of such methods, based on Monte-Carlo sampling technique, which greatly reduces the computational costs and can be applied to complex models. Moreover, using well-known results from order statistics [4], we propose to integrate the notion of numerical accuracy to our results. Our proposition mainly consists in setting some decision step before the propagation is done, rather than after it has been done. Section 2 recalls theoretical basis of the propagation technique used here and discusses previous practical solutions proposed to put this method in practice. Section 3 explains our propagating method (called the RAndom FUzzy method, or RaFu) and how it is applied. The RaFu method, implemented in SUNSET software for uncertainty analysis, is currently used and developed at IRSN

## 2. State of the art

Let us consider a set of K parameters $X_1,\ldots,X_K$ tainted with aleatory uncertainty (i.e. $X_i$ i=1,…,K takes a random value and is modelled by a probability distribution $p_i$, or equivalently by a cumulative distribution $F_i$), and a set of L parameters $X_{K+1},\ldots,X_{K+L}$ tainted with epistemic uncertainty (i.e. $X_i$ i=K+1,…,K+L has a deterministic value which is imprecisely known). Let $M(X_1,\ldots,X_K, X_{K+1},\ldots,X_{K+L})$ be the mathematical model of interest depending on our K+L uncertain parameters.

The aleatory uncertainty of a parameter X is faithfully modelled by a probability distribution p. Epistemic uncertainty, on its side, is more faithfully modelled by intervals encompassing the imprecisely known true value of a parameter. Nevertheless, we often have more information than just a minimal and a maximal values (e.g. an expert can give intervals with different confidence levels). Possibility distributions are mappings $\pi : \mathbb{R} \to [0,1]$ that can be seen as a collection of nested confidence intervals (thus extending the notion of intervals), which are the α-cuts $[x_\alpha, x^\alpha] = \{x, \pi(x) \geq \alpha\}$ of the distribution π. The degree of confidence that the interval $[x_\alpha, x^\alpha]$ contains the true value of the parameter X is then 1- α. Thus, the K first parameters of the model M are modelled by probability distributions $p_i$, while the last L are modelled by possibility distributions $\pi_i$

Parameters are then propagated through the model. Guyonnet's proposition [5] (which makes no assumption about the complexity of the model) is to first propagate the K first parameters through usual Monte-Carlo simulation, thus getting N probabilistic samples (eventually integrating some information about correlation by usual techniques [6]), and then to propagate the L last parameters by using fuzzy extension principle for each N-uple. By using the fact that the extension principle is equivalent to make an interval computation for each α-cut, he proposes to approximate the resulting fuzzy number by making computations over a limited number of α-cuts. He then gets a collection of N fuzzy numbers $^M\pi_i$, each of them occurring with probability 1/N. To each α-cut of the random fuzzy number $^M\pi$ corresponds a collection of N intervals $^M\pi_i^\alpha = [^M\pi_{i,inf}^\alpha, ^M\pi_{i,sup}^\alpha]$, from which can be built two cumulated distributions $[F_\alpha, F^\alpha]$. To build a summarized representation, Baudrit et al. [3] propose a post-processing that consists in taking the mean of the cumulated distributions $[F_\alpha, F^\alpha]$, while Ferson and Ginzburg [2] propose to take the double pair $[F_0, F^0]$ and $[F_1, F^1]$. In the two propositions, authors suppose that the fuzzy random number is built before giving one of these two representations. This supposition is computationally costly. For example, let us suppose that 100 samplings are done on the K first parameters, and that for each of them, the corresponding fuzzy number is approximated by taking twenty α-cuts (α = 0,0.05,…0.95,1). 2100 interval computations are then needed to build the final result.

In some applications, assuming one can afford so much computations is clearly unrealistic. Moreover, although it is proposed in [2] and [3,5] to use numerical sampling for complex models, the question of numerical accuracy is not considered in any of them. This is why we propose a method where numerical accuracy is integrated and where the decision step is set before the propagation (thus reducing computational cost). Let us note that the two post-processing methods mentioned above can be found back with our propagating method.

## 3. The Random Fuzzy (RaFu) method

The RaFu method uses the same theoretical framework as the one recalled in section 2. It is designed so that both epistemic and stochastic uncertainties are simultaneously sampled and propagated through the model, with the aim of building a given response. The main originality of the RaFu method is that this response is pre-defined by a triplet of parameters ($\gamma_S, \gamma_E, \gamma_A$) specified by a decision maker (DM):

- Parameter $\gamma_S$ corresponds to the statistical quantity chosen for modelling the stochastic uncertainty of the response,
- Similarly, parameter $\gamma_E$ corresponds to the fuzzy quantity used for modelling the epistemic uncertainty of the response,
- Finally, parameter $\gamma_A$ measures the desired numerical accuracy on the final result.

According to the DM values for ($\gamma_S, \gamma_E, \gamma_A$), the RaFu method then determines the minimal sample size and the nature of the required sampling to build the wished response. Number of calculations is thus reduced to its minimal number, in accordance with the DM choice. Moreover, computation cost can be easily evaluated, allowing the DM to eventually revise its choices before uncertainty propagation. For example, if the DM want to have an upper limit of the response 95% percentile, to be hyper-cautious about epistemic uncertainty (i.e. concentrate on α-cuts [$x_0, x^0$]) and to have a numerical certainty of 99% to cover the true value, he or she chooses the triplet ($\gamma_S, \gamma_E, \gamma_A$)=(0.95,0,0.99). By using results from order statistics [4] (an use often quoted as Wilks formula [7]), the RaFu method derives the minimal sampling size to satisfy the DM's choice (in our example, 90 calculations) and the nature of this sampling. Let us note that parameters ($\gamma_S, \gamma_E, \gamma_A$) are not forcefully numbers (i.e. $\gamma_E$ can be "every α-cut, from 0 to 1").

It is interesting to note that the post-processing methods proposed in [2] and [3] can both be translated in term of a decision on parameter $\gamma_E$. The Post-processing of Baudrit et al. [3] can be interpreted as "I want the mean pair of cumulated distribution taken over every confidence degree (i.e. α-cut) of epistemic uncertainty". The proposal from Ferson and Ginzburg [2] can be translated by "I want the most optimistic and the most pessimistic pair of cumulated distributions".

Let us get back to the example given in the previous section. With the RaFu method, knowing the desired final quantity before propagation allows to reduce computations from 2100 to 100 in the case of Baudrit et al. method (100 samples are made, and one random α-cut is chosen each time. This randomised α-cut insures us that we converge to the mean, without having to make the propagation for 21 α-cuts each time). In the same way, Ferson and Ginzburg's result can be obtained by reducing computations from 2100 to 200 (here, 2x100 computations are required, one set of 100 calculations for a fixed α-cut of level 0, and another one for a level of 1). Detailed algorithm and convergence proof will be provided in the full length paper.

## 4. Conclusions

Mixing fuzzy calculus with probabilistic propagation to get fuzzy random variable allow one to take into account both aleatory and epistemic uncertainties. A limitation of such methods is often the high computational complexity, which, according to us, is not always justified in practice. Thus, we propose a method (the RaFu method) that brings forward some decision step and can greatly increase numerical efficiency. The final results of usual post-processing methods can be found back with the RaFu method, as well as many other possible methods. Finally, we have proposed to add considerations about numerical accuracy in the process, an important point in sampling processes that is, to our knowledge, almost never mentioned in works trying to cope both with epistemic and aleatory uncertainties.